\pdfoutput=1

\documentclass[11pt]{article}

\usepackage[]{acl} 
\usepackage{booktabs}
\usepackage{amsmath}
\usepackage{arydshln} 
\usepackage{times}
\usepackage{latexsym}
\usepackage{CJKutf8}
\usepackage{booktabs}  
\usepackage{multirow}  
\usepackage{siunitx}   
\usepackage{graphicx}  
\usepackage{xtab}      
\usepackage{longtable}
\usepackage{color}
\usepackage{colortbl}
\usepackage{xcolor} 
\usepackage[T1]{fontenc}
\usepackage{xspace}
\newcommand{\method}{\textsc{RICo}} 
\usepackage[utf8]{inputenc}
\usepackage{xcolor}
\definecolor{tealgreen}{RGB}{0,177,0}
\definecolor{brickred}{RGB}{203,65,84}
\usepackage{microtype}

\usepackage{inconsolata}

\usepackage{graphicx}

\title{\method{}: Refined In-Context Contribution for  Automatic \\Instruction-Tuning Data Selection}

\author{\textbf{{Yixin Yang}, {Qingxiu Dong}, {Linli Yao}, {Fangwei Zhu}, {Zhifang Sui}} \\
State Key Laboratory of Multimedia Information Processing, Peking University\\
  {\tt \{yangyx,dqx,linliyao,zhufangwei2022\}@stu.pku.edu.cn, szf@pku.edu.cn}}

\begin{document}
\maketitle

\begin{abstract}
Data selection for instruction tuning is crucial for improving the performance of large language models (LLMs) while reducing training costs. In this paper, we propose \textbf{R}efined \textbf{Co}ntribution Measurement with \textbf{I}n-\textbf{Co}ntext Learning (\method{}), a novel gradient-free method that quantifies the fine-grained contribution of individual samples to both task-level and global-level model performance. \method{} enables more accurate identification of high-contribution data, leading to better instruction tuning. We further introduce a lightweight selection paradigm trained on \method{} scores, enabling scalable data selection with a strictly linear inference complexity. Extensive experiments on three LLMs across 12 benchmarks and 5 pairwise evaluation sets demonstrate the effectiveness of \method{}. Remarkably, on LLaMA3.1-8B, models trained on 15\% of \method{}-selected data outperform full datasets by 5.42\% points and exceed the best performance of widely used selection methods by 2.06\% points. We further analyze high-contribution samples selected by \method{}, which show both diverse tasks and appropriate difficulty levels, rather than just the hardest ones.\footnote{\url{https://annayang2020.github.io/ICon_Data_Selection/}}
\end{abstract}

\section{Introduction}
\begin{figure*}[t]
  \includegraphics[width=\linewidth]{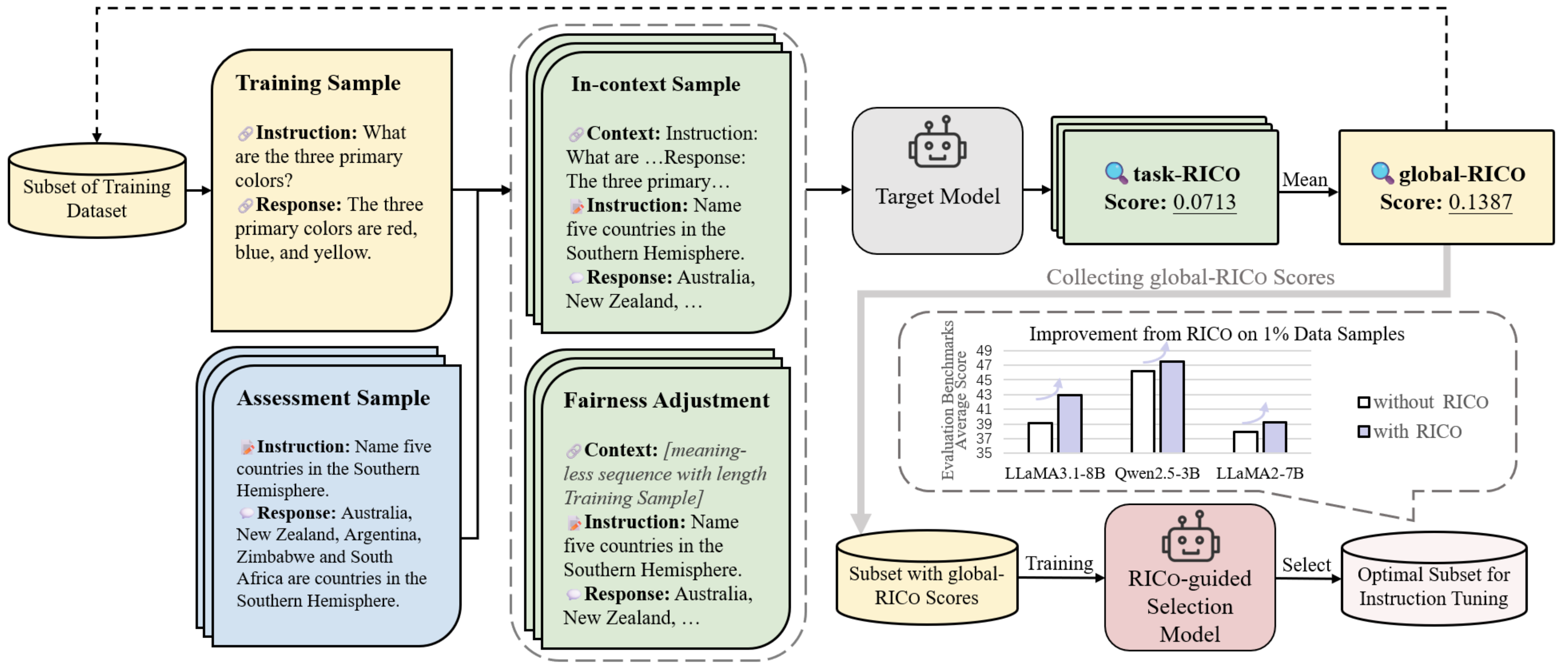}
  \caption {Overview of the \method{} Method. The diagram illustrates the two key components of the method: refined contribution quantification with the \method{} score and \method{}-guided selection paradigm training. This approach enables gradient-free, bias-reduced data selection with scalable inference.}
\label{fig:main}
\end{figure*}

Recent advancements in large language models (LLMs), such as GPT-4o~\cite{hurst2024gpt}, have demonstrated strong capabilities in both understanding and generation. These models are capable of performing a wide range of tasks~\cite{liu2023mmc, dong2023large, team2024gemma, guo2025deepseek}, often showing impressive flexibility and adaptability. Instruction tuning~\cite{mishra2021natural, Wei2021FinetunedLM, sanh2022multitask, Liu2023MMCAM, Xu2024MagpieAD} has emerged as a powerful approach to enhance the performance of such models by finetuning them to better follow human instructions. 
While traditional instruction tuning relies on amassing vast datasets~\cite{kopf2023openassistant, chung2024scaling}, recent work~\cite{Zhou2023LIMALI} shows that manual selection of high-quality data subsets can achieve  better performance with lower computational cost.

Automated data selection is crucial, as manual selection is costly and impractical. Previous studies on automated data selection~\cite{Xia2024LESSSI, Zhang2024TAGCOSTG} have explored gradient-based approaches to estimate sample value, but these are often computationally expensive.  Others propose lightweight methods based on manually designed heuristics, such as textual features~\cite{Liu2023WhatMG, Bukharin2023DataDM, Zhang2025TheBI} and instruction difficulty~\cite{Li2023FromQT, Li2024SuperfilteringWD}, but these do not directly assess training impact and can introduce human inductive bias into the selection process. Recent works~\cite{li2023one, jiao2025feasibility} adopt binary comparisons in in-context learning (ICL) settings. While the implicit fine-tuning nature of ICL~\cite{dong2022survey, dai2023can, zheng2023can, yin-etal-2024-deeper} shows promise in addressing prior limitations, existing approaches remain limited in robustly assessing sample value. They provide only coarse-grained signals, are often biased toward longer sample~\cite{wang2022perplexity}, and typically require a large number of inference calls.




In this paper, we introduce \textbf{R}efined \textbf{Co}ntribution Measurement with \textbf{I}n-\textbf{Co}ntext Learning (\method{}), a novel gredient-free selection method that captures the fine-grained contribution of individual samples to both task-level performance and overall model capability. \method{} measures each sample’s contribution across diverse tasks while reducing biases such as length sensitivity and human inductive bias. We argue that such refined measurement enables more accurate identification of high-contribution data, leading to stronger instruction tuning results. Moreover, we train a lightweight selection paradigm on these contributions, reducing inference calls and enabling scalable data selection over large candidate pools.

Our approach measures the refined contribution of each sample through three components: 1) an assessment set that provides a reliable reference for model performance, 2) \method{} scores that quantify the fine-grained contribution, and 3) a lightweight selection paradigm for scalable data selection. The \method{} score reflects the contribution of each sample to task-specific and overall performance, including a fairness adjustment to mitigate length-related bias. A higher positive score indicates a greater contribution, while a lower negative score reflects a more detrimental effect. Finally, the \method{}-guided selection paradigm is efficiently trained with LoRA~\cite{Hu2021LoRALA} on the target model, reducing selection complexity to linear inference calls with respect to the number of training samples.

We demonstrate the effectiveness of \method{} across multiple models and evaluation settings. Experiments on LLaMA3.1-8B~\cite{grattafiori2024llama}, Qwen2.5-3B~\cite{qwen2.5}, and LLaMA2-7B~\cite{lyu2024keeping} are conducted using 12 widely used benchmarks and 5 pairwise comparison test sets. Models trained on a small fraction of \method{}-selected data outperform their full-data counterparts. Notably, with only 15\% of the data, the LLaMA3.1-8B and Qwen2.5-3B models improve by 5.42 and 1.24 percentage points, respectively; LLaMA2-7B achieves a 0.65-point gain using just 5\%. \method{} also surpasses widely used selection methods. For example, \method{} improves LLaMA3.1-8B's average benchmark score by 2.06 points over the best prior method. Furthermore, we also analyze the impact of data scale, cross-dataset generalization, and the properties of high-contribution samples selected by \method{}. The main contributions of our paper are as follows:

\begin{itemize}
    \item We propose \method{}, a novel gradient-free data selection method that quantifies the refined contribution of individual samples to both task-level and overall model performance, with fairness adjustments to reduce length-related bias.
    
    \item We introduce a lightweight selection paradigm trained on \method{} contribution scores, enabling scalable data selection with a strictly linear inference complexity.

    \item We demonstrate the effectiveness of \method{} through experiments on multiple models, 12 widely used benchmarks, and 5 pairwise test sets, showing consistent performance gains over baselines. 
\end{itemize}


\section{Related Work}
\label{sec:related_work}
\paragraph{Automatic Data Selection}
Automatic data selection seeks to algorithmically construct optimal datasets to improve model performance~\cite{murphy2012machine}, reduce training cost~\cite{suarez2019asynchronous, sorscher2022beyond}, mitigate undesirable model behaviors~\cite{longpre2024pretrainer}, and ensure evaluation quality~\cite{oren2024proving}. With the rise of LLMs, it plays a central role across various training stages, including pretraining, instruction tuning, alignment, in-context learning, and task-specific finetuning~\cite{albalak2024survey}. Pretraining selection focuses on filtering large-scale raw data~\cite{soldaini2024dolma, penedo2023refinedweb}; instruction tuning will be discussed in the next section; alignment involves model-based evaluation~\cite{wang2023shepherd, cui2023ultrafeedback} and reward model re-weighting~\cite{touvron2023llama, pace2024west}; in-context learning emphasizes demonstration choice~\cite{xu2024misconfidence, luo2024context} and order~\cite{lu2021fantastically}; and task-specific tuning leverages utility-based~\cite{ivison2023data} or empirical methods~\cite{grangier2022trade}.

\paragraph{Automatic Instruction-Tuning Data Selection}
Instruction-tuning data selection aims to identify the most informative subsets of data to improve model performance during instruction tuning. Gradient-based methods~\cite{Xia2024LESSSI, Zhang2024TAGCOSTG,Pan2024GDIGTG, Chang2024TargetAwareLM} assess sample influence through gradients or subset training outcomes, but incur substantial computational cost. Heuristic-based methods~\cite{Liu2023WhatMG,Bukharin2023DataDM,Liu2024SelectITSI,Li2024SuperfilteringWD,Zhang2025TheBI} use manually crafted indicators, which may not fully capture the true training impact and can introduce human inductive bias. Appendix~\ref{sec:appendix_refs} contains the full list of heuristic-based methods cited in this work. In-context probing (ICP) methods~\cite{li2023one, jiao2025feasibility} leverage the implicit fine-tuning effect of ICL without model updates, reducing both human bias and computational cost. However, existing approaches typically rely on simple binary comparisons, yielding coarse-grained signals, exhibiting length bias, and incurring inference costs that grow multiplicatively with the size of the candidate pool. Inspired by the advantages of ICL, \method{} introduces a fine-grained, bias-reduced contribution measurement and a lightweight selection paradigm that enables scalable data selection with strictly linear inference complexity.

\section{Methodology}

To accurately and efficiently identify high-contribution samples from a training candidate set, we propose \method{}, a novel gradient-free data selection method that measures the refined contribution of individual samples. As illustrated in Figure~\ref{fig:main}, \method{} consists of three components: 1) constructing a representative assessment set, 2) computing the \method{} score to quantify fine-grained sample contribution, and 3) training a \method{}-guided selection model to curate the full dataset with strictly linear complexity. 


\subsection{Assessment Set Construction} 
\label{sec:assessment}

The goal of this component is to construct an assessment set that spans diverse tasks and reliably reflects the model’s overall capabilities. To promote reliability and mitigate bias in data origin and task design, we sample from three sources: ChatGPT-generated, GPT-4-generated, and human-authored data, resulting in a high-quality set covering various task types and instruction complexities. The assessment set is used solely for assessment, with no overlap with training data or downstream benchmarks, ensuring independence and fair comparison. Formally, we define the assessment set as $D_a$, consisting of $n$ instruction tuning samples, each represented as a pair $(x^a, y^a)$, where $x^a = \text{map}(\text{Instruction}, [\text{Input}])$ is the full instruction formed by concatenating the instruction and the optional input, and $y^a$ is the corresponding response. Meanwhile, the candidate training set is denoted as $D_t = \{(x^t_1, y^t_1), (x^t_2, y^t_2), \ldots, (x^t_m, y^t_m)\}$, where $m$ is the number of instruction-response pairs.


\subsection{Refined Contribution Quantification}
\label{sec:score}

We introduce the \method{} score, a principled metric for quantifying the refined contribution of individual samples to model performance. It enables fine-grained assessment at both task-specific and global levels, while being low-cost and bias-reduced.

\paragraph{Perplexity} Perplexity directly evaluates the likelihood of response tokens and provides a smoother measure of model performance compared to accuracy. It aligns well with instruction tuning objectives, making it a commonly adopted metric for both evaluation and data selection in instruction-tuning tasks~\cite{Li2023FromQT,Li2024SuperfilteringWD,jiang-etal-2024-instruction,li2023one}. Specifically, for a task sample $S_i = (x^a_i, y^a_i) \in D_a$, where $D_a$ is the assessment set, $y^a_i$ consists of $N$ tokens and the $k$-th token is denoted as $y^a_{i,k}$, the model’s performance under parameters $\theta$ on task $i$ is defined as shown in Equation~\ref{eq:example1}. To evaluate the contribution of a training sample $T_j = (x^t_j, y^t_j)$ from a subset $D_t' \subseteq D_t$, the sample is inserted into the context during inference. The model’s performance on $S_i$ is then reassessed after implicitly conditioning on $T_j$. The perplexity in this setting is denoted as Equation~\ref{eq:example1-1}.


{\small
\begin{equation} 
\begin{aligned}
    \label{eq:example1}
    &\text{PPL}_\theta(S_i) = \text{PPL}_\theta(y_i^a|x_i^a)\\
    & = \exp \left( - \frac{1}{N} \sum_{k=1}^{N} \log p_\theta(y^a_{i,k} | x^a_i, y^a_{i,1}, ..., y^a_{i,k-1}) \right) 
\end{aligned}
\end{equation}

\begin{equation} 
    \label{eq:example1-1}
    \text{PPL}_\theta(S_i \mid T_j) = \text{PPL}_\theta(y^a_i \mid T_j, x^a_i)
\end{equation}
}

\paragraph{Fairness Adjustment} Despite its wide use,  we observe that perplexity tends to decrease with longer inputs, even when the content is uninformative~\cite{wang2022perplexity}. This length sensitivity introduces a bias toward longer samples and fails to fairly reflect true sample contribution. To address this, we introduce a fairness adjustment to neutralize length effects. Specifically, we define the baseline performance on $S_i$ by replacing $T_j$ with a randomly generated, semantically meaningless sequence $T_{j}^{\text{rand}}$ of the same length. The length-controlled reference perplexity is defined as:

{\small
\begin{equation} 
    \label{eq:example2-0}
    \text{PPL}_\theta(S_i \mid T_j^{\text{rand}}) = \text{PPL}_\theta(y^a_i \mid T_{j}^{\text{rand}}, x^a_i)
\end{equation}
}

\paragraph{Task-level \method{} Score} We further define the task-level \method{} score to quantify the refined contribution of an individual sample on model behavior for a specific task. The score integrates the above fairness adjustment to reduce length bias. We also normalize for task difficulty to reduce its interference in contribution scoring, ensuring that the score highlights the effect of the sample rather than the task’s inherent difficulty. Specifically, we define the task-level \method{} score of sample \( S_j \) on task \( i \) under parameters \(\theta\) as shown in Equation~\ref{eq:example2}, where \( \epsilon \) is a small constant to avoid division by zero. A higher task-level \method{} score indicates a greater contribution of $S_j$ to model performance on task $i$.

{\small
\begin{equation}
\begin{aligned}
  \label{eq:example2}
  \text{task-\method{}}_\theta(T_j \xrightarrow{} S_i)  
  = \frac{\text{PPL}_\theta(S_i|T_j^{\text{rand}}) -\text{PPL}_\theta(S_i|T_j)}{\text{PPL}_\theta(S_i) + \epsilon} 
\end{aligned}
\end{equation}
}
{\small
\begin{equation}
  \label{eq:example3}
\textnormal{global-\method{}}_\theta(S_j) = \frac{1}{n} \sum_{S_i \in D_a} \textnormal{task-\method{}}_\theta(T_j \xrightarrow{} S_i)
\end{equation}
}
\paragraph{Global-level \method{} Score} To quantify a sample’s overall contribution to the model’s performance, we compute the global-level \method{} score by aggregating its task-level scores. The diversity and coverage of the assessment set provide a reliable reference for reflecting overall model performance. As we aim to improve generalization rather than specialization in any particular task type, we assign equal weight to each task. The global-level score, shown in Equation~\ref{eq:example3}, reflects the refined overall contribution of a sample and enables ranking based on total impact. Higher global-level \method{} scores indicate greater overall contribution to model performance.


\begin{figure}[t]
  \includegraphics[width=\linewidth]{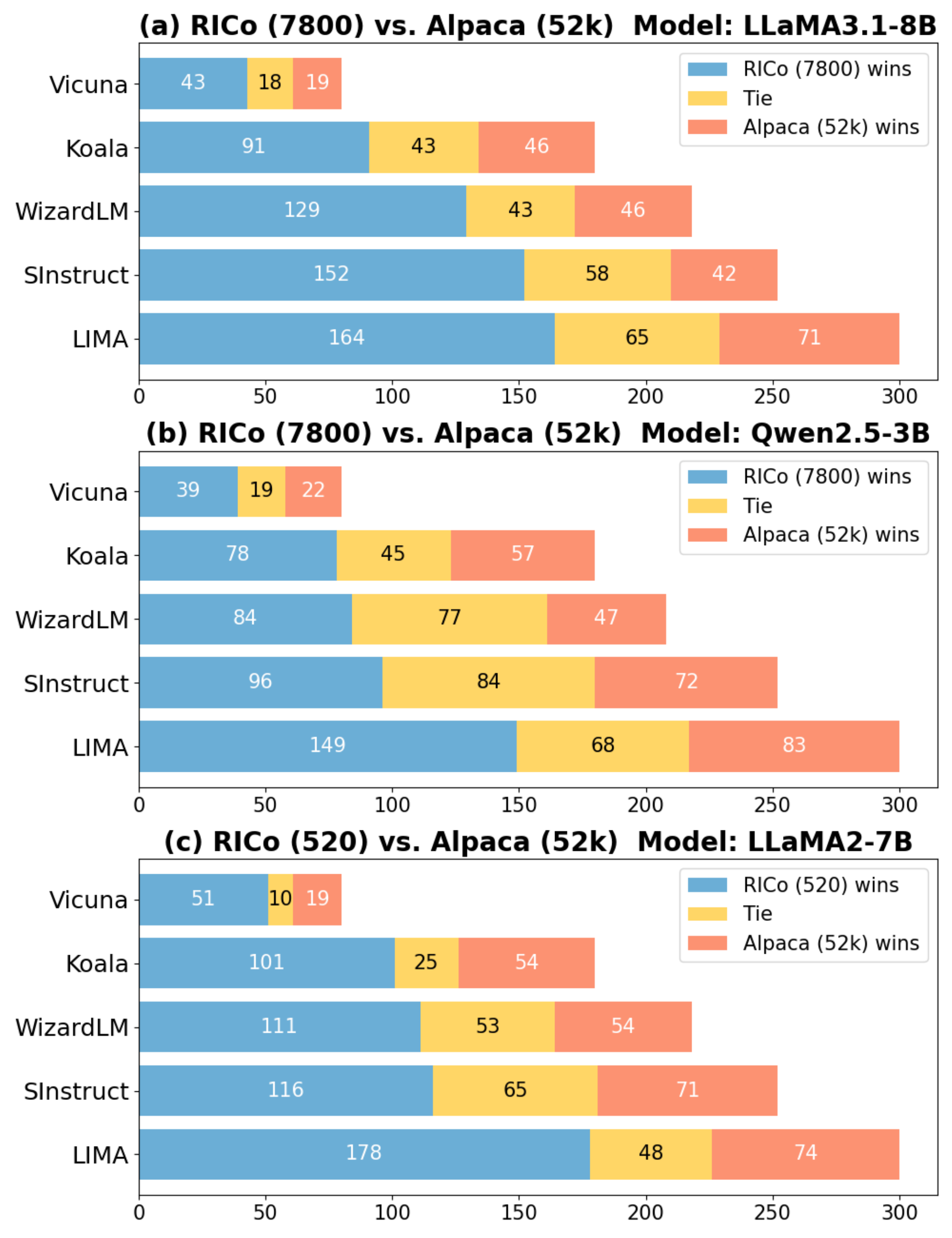}
  \caption {Pairwise comparison results (win-tie-lose) between models trained on \method{}-selected subsets and models trained on the full Alpaca dataset across five evaluation benchmarks: Vicuna, Koala, WizardLM, SInstruct, and LIMA. \method{}-selected data enhances instruction-following ability with fewer samples, as evidenced by pairwise comparisons.}
   \label{fig:compare}
\end{figure}
\subsection{\method{}-guided Selection Paradigm}
\label{sec:paradigm}

To further improve selection efficiency, we train an \method{}-guided selection model using the computed global-level \method{} scores. This reduces the selection complexity from $O(nm)$ to $O(m)$ inference calls, where $n$ is the number of assessment samples and $m$ is the number of training candidates. As a result, it enables efficient and scalable identification of high-contribution samples for instruction tuning. Specifically, we sample a subset of training data and label the top $K\%$ of samples by global-level \method{} score as high-contribution. A selection model is then trained using LoRA on the target model to classify high-contribution samples. Once trained, this selection paradigm can be applied to the full candidate pool, enabling linear-time selection in large-scale settings, with potential for practical use in realistic applications.

\begin{table*}[htbp]
    \centering
    \setlength{\tabcolsep}{3.5pt}
    \resizebox{\textwidth}{!}{
    \begin{tabular}{lccccccccccccc|c}
\toprule 
\multicolumn{1}{c}{} & \multicolumn{2}{c}{\textbf{Instruction Following}$\uparrow{}$} & \multicolumn{4}{c}{\textbf{Knowledge}$\uparrow{}$} & \multicolumn{6}{c}{\textbf{Reasoning}$\uparrow{}$} & \multirow{2}{*}{\textbf{Avg}$\uparrow{}$} & \textbf{FLOPs}$\downarrow$\\
\cmidrule(lr){2-3} \cmidrule(lr){4-7} \cmidrule(lr){8-13}
 & \textbf{IFEval} & \textbf{AlpacaEval}  & \textbf{GLUE} & \textbf{GPQA} & \textbf{MMLU} & \textbf{TQA}  & \textbf{ARC} & \textbf{BBH} & \textbf{HS} & \textbf{LQA} & \textbf{MuSR} & \textbf{WG} & & \textbf{($\times 10^{12}$)}\\

\midrule
    \rowcolor{gray!20}
    \multicolumn{15}{c}{\textbf{LLaMA3.1-8B}} \\

FULL & 15.68 & 18.66 &  54.48 & 27.27 & 43.26 & 39.60 & 40.53 & 36.46 & 51.42 & 26.73 & 37.51 & 63.77  & 37.95  & 40.12\\
\cdashline{1-15}
\method{} (1\%) & 8.31 & 28.62 &  58.47 & 31.31 & 55.77 & 43.11 &  49.91 & 44.51 & 61.19 & 23.35 & 39.37 & 71.03 & 42.91\textcolor{tealgreen}{\footnotesize (+4.96)}  & \textbf{0.59}\\
\method{} (5\%) & 16.21 & 28.96 &  56.91 & 27.27 & 53.70 & 40.51 & 47.18 & 44.23 & 58.44 & 24.42 & 37.75 & 69.46 & 42.09\textcolor{tealgreen}{\footnotesize (+4.14)}  & 3.06\\
\method{} (10\%) & 16.08 & 27.99 &  57.86 & 26.76 & 53.57 & 42.63 & 45.48 & 42.81 & 56.27 & 26.57 & 42.01 & 66.93 & 42.08\textcolor{tealgreen}{\footnotesize (+4.13)}  & 5.62\\
\method{} (15\%) & 21.16 & 29.89 &  60.58 & 30.81 & 52.30 & 43.16 & 44.80 & 44.15 & 55.56 & 28.42 & 41.64 & 67.96 & \textbf{43.37\textcolor{tealgreen}{\footnotesize (+5.42)}}  & 8.37\\
\midrule
    \rowcolor{gray!20}
    \multicolumn{15}{c}{\textbf{Qwen2.5-3B}} \\

FULL & 32.34 & 28.74  & 68.98 & 31.82 & 61.71 & 45.42 & 46.42 & 45.05 & 55.13 & 35.79 & 44.51 & 63.85 & 46.65 & 36.46 \\
\cdashline{1-15}
\method{} (1\%) & 27.04 & 34.55  & 67.22 & 31.82 & 64.63 & 44.08 & 53.16 & 47.35 & 56.19 & 34.71 & 41.05 & 68.51 & 47.53\textcolor{tealgreen}{\footnotesize (+0.88)} & \textbf{0.44}\\
\method{} (5\%) & 20.85 & 36.25 & 68.34 & 27.88 & 64.73 & 43.26 & 53.50 & 47.80 & 58.10 & 32.10 & 39.72 & 69.06 & 46.80\textcolor{tealgreen}{\footnotesize (+0.15)}  & 2.27\\
\method{} (10\%) & 25.28 & 37.05 & 69.26 & 34.85 & 63.67 & 43.35 & 50.00 & 47.00 & 57.85 & 35.79 & 40.25 & 68.27 & 47.72\textcolor{tealgreen}{\footnotesize (+1.07)} & 5.28\\
\method{} (15\%) & 23.07 & 30.11 & 70.46 & 34.34 & 64.09 & 46.90 & 51.71 & 47.68 & 58.16 & 38.40 & 41.96 & 67.80 &  \textbf{47.89\textcolor{tealgreen}{\footnotesize (+1.24)}}  & 6.85\\
\midrule
    \rowcolor{gray!20}
    \multicolumn{15}{c}{\textbf{LLaMA2-7B}} \\

FULL    & 15.48 & 21.76 & 53.98 & 26.26 & 43.69 & 40.48 & 41.72 & 36.18 & 53.98 & 25.03 & 41.11 & 62.82 & 38.54 & 48.72 \\
\cdashline{1-15}
\method{} (1\%)  & 5.25 & 30.35 & 54.51 & 23.23 & 40.55 & 45.82 & 42.24 & 38.34 & 59.20 & 25.19 & 37.68 & 67.96 &  \textbf{39.19\textcolor{tealgreen}{\footnotesize (+0.65)}} & \textbf{0.66} \\
\method{} (5\%)  & 7.78 & 31.47 & 51.93 & 23.23 & 36.95 & 45.72 & 40.53 & 36.07 & 57.57 & 23.96 & 40.62 & 66.77 & 38.55\textcolor{tealgreen}{\footnotesize (+0.01)} & 3.26 \\
\method{} (10\%)  & 11.86 & 29.76 & 50.62 & 25.76 & 37.30 & 45.19 & 41.72 & 35.38 & 57.39 & 24.88 & 39.82 & 66.38 & 38.84\textcolor{tealgreen}{\footnotesize (+0.30)} & 6.16\\
\method{} (15\%) & 8.06 & 26.83 & 54.67 & 27.78 & 36.39 & 43.89 & 42.66 & 34.74 & 57.01 & 23.50 & 39.80 & 66.06 & 38.45\textcolor{brickred}{\footnotesize (-0.09)} & 9.18 \\
\bottomrule
\end{tabular}
    }
    \caption{Performance of LLaMA3.1-8B, Qwen2.5-3B, and LLaMA2-7B in terms of FLOPs and evaluation benchmarks for instruction following, knowledge, and reasoning. The datasets TQA, HS, LQA, and WG correspond to TruthfulQA, HellaSwag, LogicQA, and WinoGrande, respectively. Models trained on \method{}-selected data outperform those trained on full datasets on evaluation benchmarks, achieving better performance with fewer samples.}
    \label{tab:benchmarks}
\end{table*}
\section{Experiments Setup}

\subsection{Datasets}

\paragraph{Training Dataset} 

In this paper, we use the classic Alpaca dataset~\cite{taori2023stanford}, one of the original instruction datasets created by Stanford University. The Alpaca dataset consists of 52,002 instruction-following samples, generated with TextDavinci-003 through the self-instruction approach~\cite{wang2022self}. The WizardLM dataset~\cite{xu2023wizardlm} utilizes the Evol-Instruct algorithm to enhance the quality and complexity of instruction data. We conduct experiments using WizardLM-70K to verify the generalization of our method across different datasets.

\paragraph{Assessment Dataset} 
We consider the impact of a training candidate sample on the model’s performance over the assessment dataset as an indicator of its overall influence on model capability, as constructed in Section~\ref{sec:assessment}. To reduce bias and ensure diversity, we sample data from three sources: OpenOrca-GPT3.5~\cite{lian2023openorca} containing ChatGPT-generated data, OpenOrca-GPT4 containing GPT-4-generated data, and Dolly-15K~\cite{mike2023free} containing human-generated data. From these three datasets, we randomly select 1,020 instructions to construct the assessment set. 


\subsection{Implementation Details}




For experiments on the LLaMA3.1-8B~\cite{grattafiori2024llama}, Qwen2.5-3B~\cite{qwen2.5}, and LLaMA2-7B~\cite{lyu2024keeping}, we all follow the original Alpaca\footnote{\url{https://github.com/tatsu-lab/stanford_alpaca}} training configuration using its codebase. Models are trained for three epochs with Adam optimizer~\cite{kingma2014adam}, a learning rate of \(2 \times 10^{-5}\), and a batch size of 128. The maximum input length is set to 512 for Alpaca dataset and 2048 for WizardLM dataset.
 
\subsection{Evaluation Metrics}

\subsubsection{Benchmark Evaluation}

To evaluate the effectiveness of \method{}, we report performance on several widely used benchmarks. These benchmarks fall into three categories, assessing different capabilities: instruction following, knowledge, and reasoning. Instruction following evaluation includes: IFEval~\cite{zhou2023instruction} and AlpacaEval~\cite{alpaca_eval}. AlpacaEval provides automatic evaluation on the AlpacaFarm set~\cite{dubois2023alpacafarm}, where we use GPT-4~\cite{achiam2023gpt} as both the response generator and evaluator. Knowledge evaluation includes: GLUE~\cite{wang2018glue}, GPQA~\cite{rein2024gpqa}, MMLU~\cite{hendrycks2020measuring}, and TruthfulQA~\cite{lin2021truthfulqa}. Reasoning evaluation includes: ARC~\cite{clark2018think}, BBH~\cite{suzgun2022challenging},  HellaSwag~\cite{zellers2019hellaswag}, LogiQA~\cite{liu2020logiqa}, MuSR~\cite{sprague2023musr}, and Winogrande~\cite{sakaguchi2021winogrande}.

\subsubsection{Pairwise Comparison}
The performance on pairwise comparison test sets for instruction following is also provided, including WizardLM~\cite{xu2023wizardlm}, Self-Instruct (SInstruct) ~\cite{wang2022self}, Vicuna~\cite{chiang2023vicuna}, Koala~\cite{vu2023koala}, and LIMA~\cite{zhou2023lima}. These sets contain 218, 252, 80 and 300 human-curated instruction data, respectively, across domains such as math, coding, writing, and general knowledge. We use GPT-4~\cite{hurst2024gpt} as the judge. For each instruction, candidate models generate responses, which are scored by GPT-4 in a pairwise fashion on a scale of 1 to 10 based on relevance, accuracy, and other criteria. To mitigate positional bias~\cite{wang2023large}, we submit each response pair twice with reversed order. A model is considered to win only if it does not lose in both orders, following these rules: \textbf{Win}: wins both, or wins one and ties the other; \textbf{Tie}: ties both, or wins one and loses one; \textbf{Lose}: loses both, or loses one and ties the other.

\section{Experiment Results}
\subsection{Main Results}

In this section, we present the evaluation benchmark results shown in Table~\ref{tab:benchmarks}. The models are trained using different proportions of Alpaca data selected by the \method{} method, including 1\%, 5\%, 10\%, and 15\%, corresponding to 520, 2,600, 5,200, and 7,800 samples, respectively. FULL denotes models trained on the full, unfiltered dataset. Experiments are conducted on multiple models, including LLaMA3.1-8B, Qwen2.5-3B, and LLaMA2-7B. Models trained on a small fraction of \method{}-selected data often match or exceed the performance of full-data models across all benchmarks, with consistent gains in average benchmark scores. Specifically, the LLaMA3.1-8B model trained on just 15\% of \method{}-selected data exceeds the full-data baseline by 5.42 percentage points. Similarly, the Qwen2.5-3B model with 15\% \method{}-selected data and the LLaMA2-7B model with only 1\% achieve overall gains of 1.24\% and 0.65\%, respectively. Notably, these \method{}-trained models require only about 1/5 to 1/100 of the FLOPs compared to full-data training, demonstrating significantly lower resource consumption. These results underscore the effectiveness of \method{} in enhancing instruction tuning with substantially less data.


\begin{table}[t]
\centering
\small
\setlength{\tabcolsep}{3.2mm}{
\begin{tabular}{l|cccc}
\toprule
 & IF & KN & RS & Avg\\ \midrule
Random & 12.13 & 45.18 & 46.89 & 40.52 \\
Low PPL & 19.35 & 44.45 & 46.53 & 41.31 \\
Top PPL & 3.28 & 41.72 & 44.69 & 36.80\\ 
Alpagasus & 16.39 & 45.14 & 45.29 & 40.42 \\ 
Deita &15.20 & 45.46 & 45.92 & 40.65 \\ 
Superfilter & 20.25 & 44.76 & 45.45 & 41.02 \\ 
Nuggets & 18.63 & 44.56 & 46.22 & 41.07 \\
\midrule
\method{} (Ours) & \textbf{25.52} & \textbf{46.71} & \textbf{47.09} & \textbf{43.37} \\ 
\bottomrule
\end{tabular}}
\caption{Performance of different methods on evaluation benchmarks. The `IF', `KN', and `RS' correspond to average scores on Instruction Following, Knowledge, and Reasoning, respectively. The models are trained on LLaMA3.1-8B with 15\% selected samples. \method{} outperforms widely used data selection methods on evaluation benchmarks.}
\label{tab:statistics}
\end{table}


The results of the pairwise evaluation, including detailed win-tie-lose statistics on the Vicuna, Koala, WizardLM, SInstruct, and LIMA test sets, are presented in Figure~\ref{fig:compare}. We focus on the best-performing models from the evaluation benchmarks: LLaMA3.1-8B, Qwen2.5-3B, and LLaMA2-7B, each trained on 15\%, 15\%, and 1\% \method{}-selected samples, respectively. All three models consistently outperform their counterparts trained on the full Alpaca dataset across all five test sets. These results further validate the effectiveness of \method{} in improving instruction-following ability, even with significantly less training data.
\subsection{Comparison with Other Methods}

\begin{table*}[h]
    \centering
    \resizebox{0.85\textwidth}{!}{
    \begin{tabular}{l|ccccc|c}
        \hline
         \textbf{Comparison}  & \multicolumn{5}{c|}{ \textbf{Pairwise Winning Score $\uparrow$}} & \multirow{2}{*}{\textbf{Overall}} \\
        Test Set & Vicuna & WizardLM & LIMA & SInstruct  & Koala \\
        \hline
        \method{}  vs. Random & 1.2125 & 1.4541 & 1.3233 & 1.5159 & 1.3389 & 1.3922 \\
        \method{} vs. Low PPL & 1.3375 & 1.3945 & 1.4533 & 1.3690 & 1.3333 & 1.3903 \\
        \method{} vs. Top PPL  & 1.9125 & 1.8073 & 1.8567 & 1.8452 & 1.7778 & 1.8340 \\
        \method{} vs. Alpagasus~\cite{Chen2023AlpaGasusTA} & 1.1250 & 1.1789 & 1.1033 & 1.3611 & 1.3389 & 1.2252 \\
        \method{} vs. Deita~\cite{Liu2023WhatMG}  & 1.3375 & 1.2661 & 1.2733 & 1.3730 & 1.3889 & 1.3214 \\
        \method{} vs. Superfilter~\cite{Li2024SuperfilteringWD}  & 1.0375 & 1.0780 & 1.0100 & 1.1627 & 1.0944 & 1.0786 \\
        \method{} vs. Nuggets~\cite{li2023one}  & 1.1625 & 1.1422 & 1.0967 & 1.1508 & 1.1111 & 1.1272 \\
        \hline
    \end{tabular}
}
    \caption{Comparison with other methods on Vicuna, WizardLM, LIMA, SInstruct, and Koala test set. The pairwise Winning Scores are calculated between models using our method and other methods. All the comparisons are performed by GPT-4, and the values that are greater than 1.0 represent our models are better and vice versa. The models are trained on LLaMA3.1-8B with 15\% selected samples. \method{} consistently outperforms widely used data selection methods on pairwise comparison test sets.} 
    \label{tab:comparison}
\end{table*}

In this section, we compare our method with several widely used data selection baselines on the Alpaca dataset, as shown in Table~\ref{tab:comparison} and Table~\ref{tab:statistics}. \textit{Random} selects samples uniformly at random. \textit{Low-PPL} and \textit{Top-PPL} select samples with the lowest and highest perplexity scores, respectively. \textit{Alpagasus}~\cite{Chen2023AlpaGasusTA} uses GPT-3.5-Turbo to score responses based on helpfulness, accuracy, and other dimensions. \textit{Deita}~\cite{Liu2023WhatMG} leverages models fine-tuned from LLaMA and Mistral to assess sample quality, complexity, and diversity. \textit{Superfilter}~\cite{Li2024SuperfilteringWD} ranks samples using IFD scores derived from GPT-2 to estimate instruction difficulty, assuming that smaller models provide sufficiently reliable signals. \textit{Nuggets}~\cite{li2023one}, a representative ICP-based method, uses binary comparisons in the ICL setting to assign golden scores. We adopt its KMeans\textsubscript{100} selection method, consistent with its official implementation.

Table~\ref{tab:comparison} presents the pairwise winning scores of the \method{} model against other data selection baselines. The model is trained on 15\% \method{}-selected samples using LLaMA3.1-8B as the base model. The pairwise winning score is calculated as \((\text{Num(Win)} - \text{Num(Lose)})/\text{Num(All)} + 1\), providing a direct performance comparison between \method{} and each baseline. A score above 1 indicates that \method{} outperforms the corresponding method, with larger values reflecting stronger advantages. \method{} consistently outperforms all baselines across the five test sets and in overall comparison. Table~\ref{tab:statistics} presents results on 13 evaluation benchmarks, reporting average scores in instruction tuning, knowledge, reasoning, and overall performance. \method{} achieves the best performance across all categories, demonstrating its effectiveness in selecting high-contribution data for instruction tuning. Detailed results are provided in Appendix~\ref{sec:appendix_A}.


\section{Analysis}

\begin{figure}[t]
  \includegraphics[width=\linewidth]{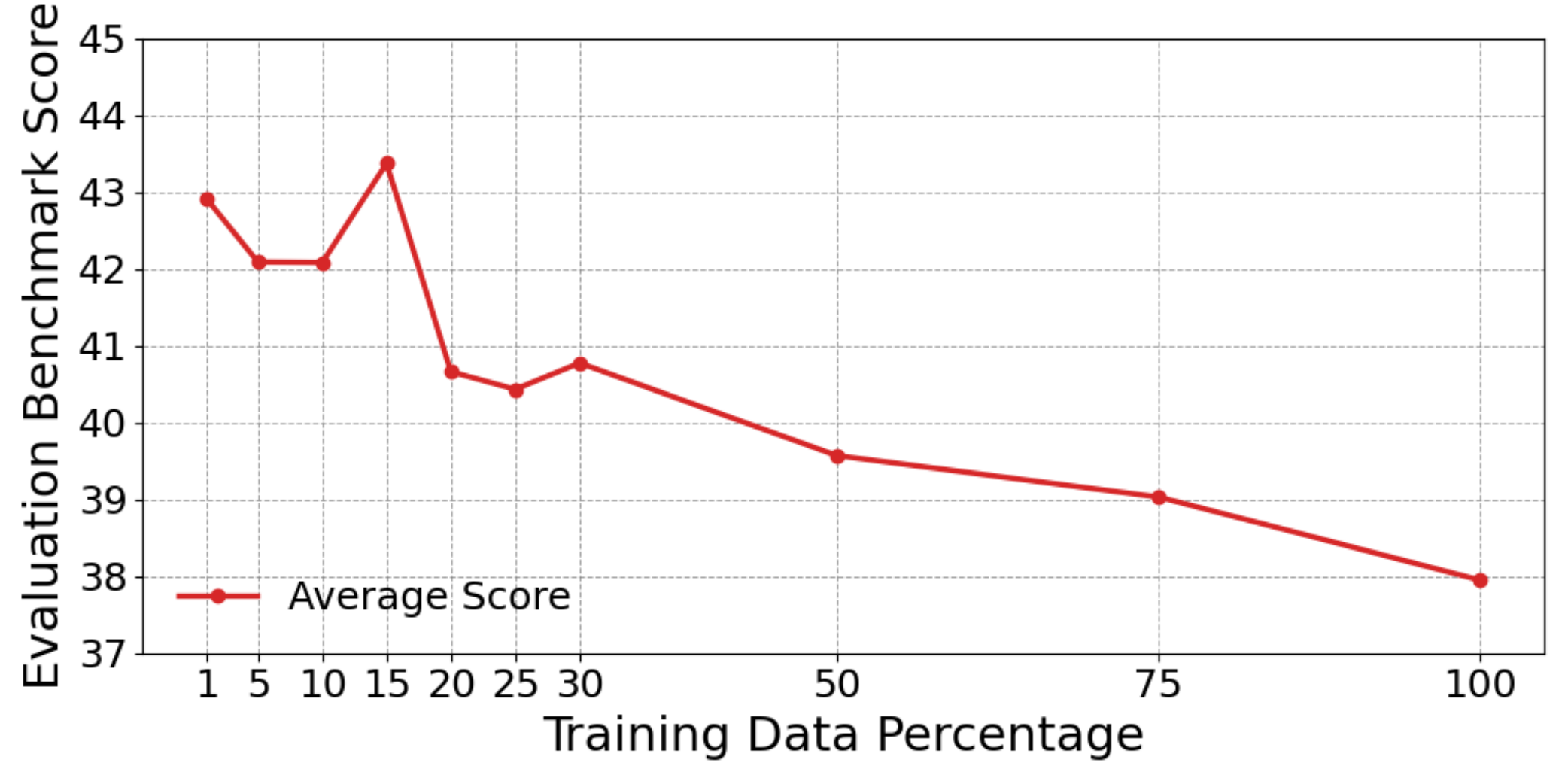}
  \caption {Model performance (average score) on evaluation benchmarks with varying proportions of \method{}-selected Alpaca data. The models are trained based on LLaMA3.1-8B. The dataset is Alpaca dataset. As data scale grows, performance generally improves and then declines, with the best result at 15\%.}
   \label{fig:data_scale}
\end{figure}

\paragraph{Optimal Data Scale for Instruction Tuning}
We further investigate how the scale of selected data affects model performance. Using the LLaMA3.1-8B model as a case study, Figure~\ref{fig:data_scale} presents the results of training on \method{}-selected subsets of the Alpaca dataset at varying proportions: 1\%, 5\%, 10\%, 15\%, 20\%, 25\%, 30\%, 50\%, 75\%, and 100\%. The average score across all evaluation benchmarks generally rises and then declines, peaking at 15\%, indicating that there exists an optimal data scale for instruction tuning. The performance drop beyond this point suggests that low-contribution samples begin to dilute the training signal, underscoring the effectiveness of \method{} in prioritizing high-contribution data. Based on this observation, we select the model trained on 15\% of \method{}-selected data as the representative model for comparisons. Detailed results for each selection scale are provided in Appendix~\ref{sec:appendix_B}.

\paragraph{Generalization of \method{} on different Datasets}
\label{tab:statistics_w}

\begin{table}[t]
\centering
\small
\setlength{\tabcolsep}{3.0mm}{
\begin{tabular}{l|cccc}
\toprule
 & IF & KN & RS & Avg\\ \midrule
FULL  & 33.61 & 45.46 & 44.37 & 42.94 \\
\cdashline{1-5}
\noalign{\vskip 0.5ex}
\method{} (1\%) & 34.57 & 46.39 & \textbf{49.57} & 46.01 \\
\method{} (5\%) & \textbf{43.21} & \textbf{48.28} & 48.36 & \textbf{47.48}\\ 
\method{} (10\%) & 36.89 & 48.15 & 47.18 & 45.79 \\ 
\method{} (15\%) &35.24 & 46.29 & 45.76 & 44.18 \\ 
\bottomrule
\end{tabular}}
\caption{Evaluation benchmark results of models trained on WizardLM dataset and its \method{}-selected subsets. The `IF', `KN', and `RS' correspond to average scores on Instruction Following, Knowledge, and Reasoning, respectively. Results on WizardLM dataset confirm the generalization of \method{} across datasets.}
\end{table}

In addition to the main experiments on the Alpaca dataset, we evaluate the generalization of \method{} to a different dataset. Specifically, we apply the selection paradigm trained on Alpaca samples to directly filter the WizardLM dataset. The results, shown in Table~\ref{tab:statistics_w} and detailed in Appendix~\ref{sec:appendix_C}, cover selection scales of 1\%, 5\%, 10\%, 15\%, and 100\%, corresponding to 700, 3,500, 7,000, 10,500, and 70K samples, respectively. As shown, models trained on \method{}-selected subsets consistently outperform those trained on the full dataset across Instruction Following, Knowledge, and Reasoning tasks, achieving higher average and overall scores. Notably, the model trained on just 5\% of the data achieves the best results, with a 4.54 percentage point improvement in average evaluation benchmark score. These findings highlight the effectiveness of \method{} across datasets. Importantly, the selection paradigm used here is directly transferred from Alpaca without modification, demonstrating its robust cross-dataset generalization.

\paragraph{Characteristics of High-Contribution \method{} Samples} 
\begin{figure}[t]
\centering
  \includegraphics[width=0.80\linewidth]{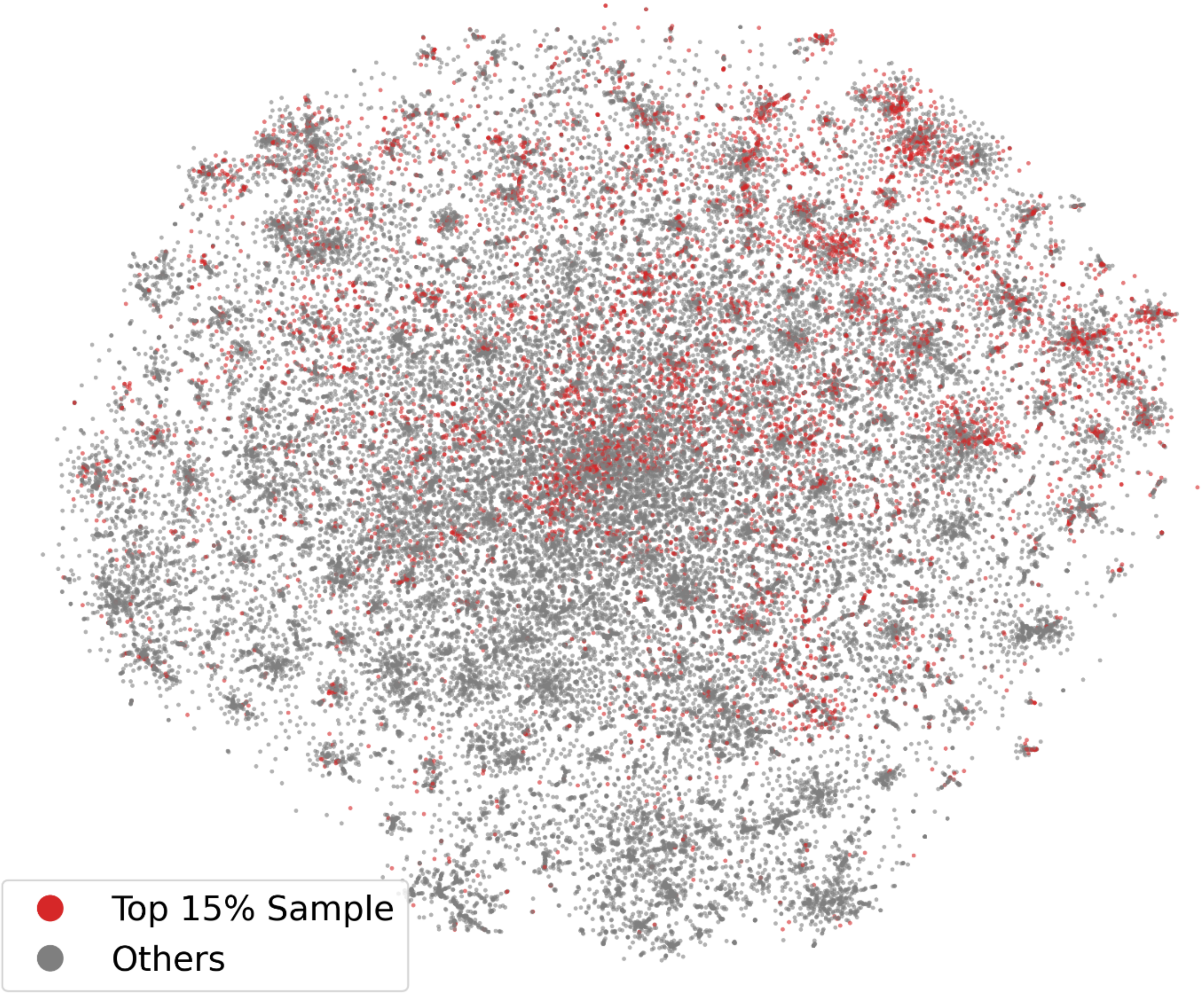}
  \caption {Visualization using t-SNE on sample embeddings from the Alpaca dataset.  Red points represent samples with the top 15\% \method{} high-contribution scores and gray points represent other samples from the dataset. High-contribution \method{} samples exhibit diverse characteristics.}
   \label{fig:Figure_4}
\end{figure}
\begin{figure}[t]
  \includegraphics[width=\linewidth]{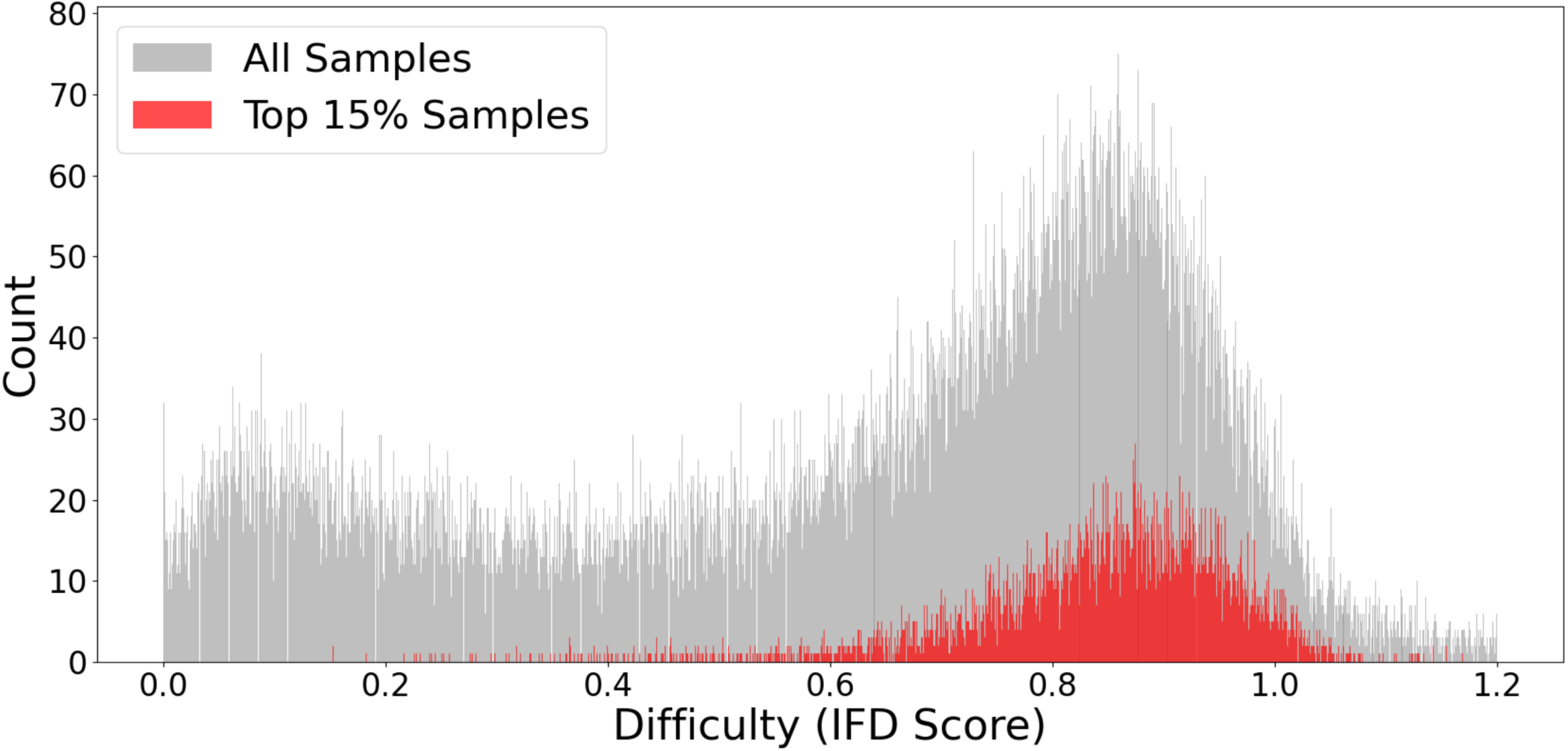}
  \caption {Difficulty distribution of top 15\% high-contribution samples vs. full Alpaca dataset. High-contribution samples (red bars), selected via \method{} for instruction tuning, exhibit distinct difficulty patterns compared to the full dataset (gray bars). High-contribution \method{} samples tend to fall within an appropriate difficulty range.}
   \label{fig:Figure_5}
\end{figure}

We further analyzed the characteristics of the high-contribution samples selected by the \method{} method. Figure~\ref{fig:Figure_4} shows the distribution of high-contribution samples in the t-SNE space, where we observe both cross-cluster dispersion and local density. This pattern aligns with the notion in previous studies~\cite{Liu2023WhatMG, Bukharin2023DataDM} that "instruction data should be diverse". However, the distribution is not uniformly scattered. Instead, some clusters exhibit higher density, suggesting that high-contribution samples tend to exhibit certain clustering behavior in the feature space. Figure~\ref{fig:Figure_5} further compares the distribution of instruction difficulty between high-contribution samples and the full dataset. We use the IFD score~\cite{Li2023FromQT, Li2024SuperfilteringWD} as the difficulty metric. The t-test~\cite{Kendall1937StatisticalMF} result (t = 2.24, p = 0.0251) shows that high-contribution samples are significantly more difficult than the full dataset on average, indicating that the selection method tends to favor more challenging samples. However, as seen in the distribution, high-contribution samples are not exclusively concentrated in the highest difficulty range. Interestingly, while the method favors more difficult samples, some mid-level difficulty samples are retained, and extreme difficulty samples are avoided. This finding challenges the view that more difficult samples are always inherently more beneficial for model training~\cite{Du2023MoDSMD, Li2023FromQT, Li2024SuperfilteringWD, Li2024SelectiveRS, Zhang2025TheBI}.

\section{Conclusion}
In this paper, we propose \method{}, a novel gradient-free data selection method that quantifies the refined contribution of individual samples for instruction tuning. \method{} accurately identifies high-contribution samples by computing task- and global-level scores with reduced bias. We further introduce a lightweight selection paradigm, enabling efficient and scalable data curation. Experiments on three LLMs across 12 benchmarks and 5 pairwise evaluation sets show that models trained on \method{}-selected data outperform full-data baselines, achieving better performance with fewer samples. \method{} also outperforms other widely used selection methods. We additionally analyze optimal data scales, cross-dataset generalization, and the properties of selected samples.

\section*{Limitations}
In estimating refined contribution, we leverage the implicit fine-tuning effect of in-context learning (ICL). Since ICL does not replicate the exact dynamics of full batch training with gradient updates, it may not fully capture all training interactions. However, it provides a lightweight and effective approximation of how individual samples influence model behavior, making it a practical tool for guiding data selection in instruction tuning. Furthermore, while our method demonstrates promising results for instruction tuning, we believe it has the potential for broader applicability. For instance, \method{} could be adapted to other tasks and domains by modifying the assessment set and the calculation of the \method{} score. Exploring these extensions is beyond the scope of this paper and remains a direction for future work.


\bibliography{custom}

\appendix

\section{Detailed Evaluation Benchmark for Different Methods}
\label{sec:appendix_A}
The detailed results for different methods on evaluation benchmarks are shown in Table~\ref{tab:detailed_mothods}. The methods include Low PPL, Top PPL, Alpagasus, Deita, Superfilter, and \method{}. Overall, \method{} outperforms existing methods across the evaluated benchmarks.
\begin{table*}[htbp]
    \centering
    \setlength{\tabcolsep}{3.5pt}
    \resizebox{\textwidth}{!}{
    \begin{tabular}{lcccccccccccccccc}
\toprule 
\multicolumn{1}{c}{} & \multicolumn{3}{c}{\textbf{Instruction Following}$\uparrow{}$} & \multicolumn{5}{c}{\textbf{Knowledge}$\uparrow{}$} & \multicolumn{7}{c}{\textbf{Reasoning}$\uparrow{}$} & \multirow{2}{*}{\textbf{Avg}$\uparrow{}$} \\
\cmidrule(lr){2-4} \cmidrule(lr){5-9} \cmidrule(lr){10-16}
 & \textbf{IFEval} & \textbf{AE} & \textbf{Avg} & \textbf{GLUE} & \textbf{GPQA} & \textbf{MMLU} & \textbf{TQA} & \textbf{Avg} & \textbf{ARC} & \textbf{BBH} & \textbf{HS} & \textbf{LQA} & \textbf{MuSR} & \textbf{WG} & \textbf{Avg} & \\

\midrule

Random & 4.85 & 19.40 & 12.13 & 55.60 & 32.80 & 51.40 & 40.90 & 45.18 & 43.69 & 40.13 & 55.40 & 31.00 & 43.20 & 67.90 & 46.89 & 40.52 \\
Low PPL & 21.28 & 17.42 & 19.35 & 56.51 & 29.80 & 53.18 & 38.30 & 44.45 & \textbf{44.97} & 40.51 & 56.32 & 30.88 & 37.37 & 69.14 & 46.53 & 41.31\\

Top PPL & 0.97 & 5.60 & 3.28 & 55.29 & 26.77 & 44.82 & 39.98 & 41.72 & 43.69 & 37.01 & 54.55 & 23.04 & 44.10 & 65.75 & 44.69 & 36.80 \\
Alpagasus & 9.72 & 23.07 & 16.39 & 57.52 & 30.81 & 53.82 & 38.39 & 45.14 & 44.20 & 41.04 & 56.13 & 27.19 & 37.74 & 65.43 & 45.29 & 40.42 \\
Deita & 9.10 & 21.30 & 15.20 & 58.31 & 27.28 & 54.47 & 41.77 & 45.46 & 42.66 & 42.94 & 55.67 & 26.57 & 39.90 & 67.79 & 45.92 & 40.65 \\
Superfilter & 13.98 & 26.53 & 20.25 & 55.64 & 27.78 & 53.66 & 41.94 & 44.76 & 45.65 & 41.31 & 56.56 & 25.81 & 37.12 & 66.22 & 45.45 & 41.02 \\
Nuggets & 13.78 & 23.48 & 18.63 & 53.95 & 30.81 & 51.69 & 41.80 & 44.56 & 43.09 & 44.18 & 54.92 & 27.50 & 40.28 & 67.32 & 46.22 & 41.07 \\
\midrule
\method{} (Ours) & 21.16 & 29.89 & \textbf{25.52} & 60.58 & 30.81 & 52.30 & 43.16 & \textbf{46.71} & 44.80 & 44.15 & 55.56 & 28.42 & 41.64 & 67.96 & \textbf{47.09} & \textbf{43.37}\\
\bottomrule
\end{tabular}
    }
    \caption{The detailed results of different methods on the evaluation benchmarks. The datasets AE, TQA, HS, LQA, and WG correspond to AlpacaEval, TruthfulQA, HellaSwag, LogicQA, and WinoGrande, respectively. All models are trained on LLaMA3.1-8B with 15\% selected samples.}
    \label{tab:detailed_mothods}
\end{table*}

\section{Detailed Evaluation Benchmark for Different Data Scales}
\label{sec:appendix_B}
The detailed results for different data scales on evaluation benchmarks are shown in Table~\ref{tab:detailed_data_scales}. As data scale increases, performance improves initially but later declines, indicating an optimal data size.

\begin{table*}[htbp]
    \centering
    \setlength{\tabcolsep}{3.5pt}
    \resizebox{\textwidth}{!}{
    \begin{tabular}{lcccccccccccccccc}
\toprule 
\multicolumn{1}{c}{} & \multicolumn{3}{c}{\textbf{Instruction Following}$\uparrow{}$} & \multicolumn{5}{c}{\textbf{Knowledge}$\uparrow{}$} & \multicolumn{7}{c}{\textbf{Reasoning}$\uparrow{}$} & \multirow{2}{*}{\textbf{Avg}$\uparrow{}$} \\
\cmidrule(lr){2-4} \cmidrule(lr){5-9} \cmidrule(lr){10-16}
 & \textbf{IFEval} & \textbf{AE} & \textbf{Avg} & \textbf{GLUE} & \textbf{GPQA} & \textbf{MMLU} & \textbf{TQA} & \textbf{Avg} & \textbf{ARC} & \textbf{BBH} & \textbf{HS} & \textbf{LQA} & \textbf{MuSR} & \textbf{WG} & \textbf{Avg} & \\

\midrule

\method{} (1\%) & 8.31 & 28.62 & 18.46 & 58.47 & 31.31 & 55.77 & 43.11 & \textbf{47.17} & 49.91 & 44.51 & 61.19 & 23.35 & 39.37 & 71.03 & \textbf{48.23} & 42.91 \\

\method{} (5\%) & 16.21 & 28.96 & 22.58 & 56.91 & 27.27 & 53.70 & 40.51 & 44.60 & 47.18 & 44.23 & 58.44 & 24.42 & 37.75 & 69.46 & 46.91 & 42.09  \\

\method{} (10\%) & 16.08 & 27.99 & 22.03 & 57.86 & 26.76 & 53.57 & 42.63 & 45.21 & 45.48 & 42.81 & 56.27 & 26.57 & \textbf{42.01} & 66.93 & 46.68 & 42.08  \\

\method{} (15\%) & 21.16 & 29.89 & \textbf{25.52} & 60.58 & 30.81 & 52.30 & 43.16 & 46.71 & 44.80 & 44.15 & 55.56 & 28.42 & 41.64 & 67.96 & 47.09 & \textbf{43.37}  \\

\method{} (20\%) & 11.95 & 27.24 & 19.60 & 55.07 & 29.29 & 53.02 & 41.71 & 44.77 & 43.36 & 42.08 & 55.06 & 26.88 & 37.36 & 64.88 & 44.94 & 40.66 \\

\method{} (25\%) & 15.63 & 23.35 & 19.49 & 50.84 & 28.28 & 51.12 & 41.26 & 42.88 & 42.06 & 42.13 & 54.52 & 28.88 & 39.77 & 67.32 & 45.78 & 40.43 \\

\method{} (30\%) & 17.72 & 25.87 & 21.79 & 58.57 & 24.74 & 52.41 & 40.66 & 44.10 & 41.89 & 42.32 & 54.35 & 26.57 & 37.63 & 66.54 & 44.88 & 40.77 \\

\method{} (50\%) & 15.00 & 22.91 & 18.95 & 57.75 & 23.73 & 51.50 & 41.12 & 43.53 & 41.64 & 40.33 & 53.19 & 28.12 & 35.27 & 64.25 & 43.80 & 39.57 \\

\method{} (75\%) & 15.17 & 23.79 & 19.48 & 55.11 & 29.80 & 46.70 & 39.94 & 42.89 & 39.99 & 37.24 & 52.52 & 28.73 & 35.12 & 64.25 & 42.98 & 39.03 \\

FULL (100\%) & 15.68 & 18.66 & 17.17 & 54.48 & 27.27 & 43.26 & 39.60 & 41.15 & 40.53 & 36.46 & 51.42 & 26.73 & 37.51 & 63.77 & 42.74 & 37.95 \\

\bottomrule
\end{tabular}
    }
    \caption{The detailed results of different data scales on the evaluation benchmarks. The datasets AE, TQA, HS, LQA, and WG correspond to AlpacaEval, TruthfulQA, HellaSwag, LogicQA, and WinoGrande, respectively. All models are trained on LLaMA3.1-8B with 15\% selected samples.}
    \label{tab:detailed_data_scales}
\end{table*}

\section{Detailed Evaluation Benchmark for Different Dataset}
\label{sec:appendix_C}
The detailed results of WizardLM dataset on evaluation benchmarks are shown in Table~\ref{tab:wizardlm123}. As demonstrated by the experimental data, \method{} exhibits strong generalization capabilities across diverse datasets.

\begin{table*}[htbp]
    \centering
    \setlength{\tabcolsep}{3.5pt}
    \resizebox{\textwidth}{!}{
    \begin{tabular}{lcccccccccccccccc|c}
\toprule 
\multicolumn{1}{c}{} & \multicolumn{3}{c}{\textbf{Instruction Following}$\uparrow{}$} & \multicolumn{5}{c}{\textbf{Knowledge}$\uparrow{}$} & \multicolumn{7}{c}{\textbf{Reasoning}$\uparrow{}$} & \multirow{2}{*}{\textbf{Avg}$\uparrow{}$} & \textbf{FLOPs}$\downarrow$\\
\cmidrule(lr){2-4} \cmidrule(lr){5-9} \cmidrule(lr){10-16}
 & \textbf{IFEval} & \textbf{AE} & \textbf{Avg} & \textbf{GLUE} & \textbf{GPQA} & \textbf{MMLU} & \textbf{TQA} & \textbf{Avg} & \textbf{ARC} & \textbf{BBH} & \textbf{HS} & \textbf{LQA} & \textbf{MuSR} & \textbf{WG} & \textbf{Avg} & & \textbf{($\times 10^{12}$)}\\

\midrule
ALL & 25.43 & 41.79 & 33.61 & 61.13 & 25.76 & 48.05 & 46.91 & 45.46 & 39.76 & 40.56 & 51.22 & 27.65 & 42.41 & 64.64 & 44.37 & 42.94 & 191.95\\
\cdashline{1-18}

\method{} (1\%) & 20.70 & 48.44 & 34.57 & 55.41 & 27.78 & 60.48 & 41.87 & 46.39 & 50.17 & 46.61 & 60.08 & 27.34 & 41.24 & 71.98 & \textbf{49.57} & 46.01 & 4.85 \\

\method{} (5\%) & 26.90 & 59.52 & \textbf{43.21} & 62.78 & 26.26 & 56.64 & 47.44 & \textbf{48.28} & 48.29 & 44.23 & 57.73 & 27.50 & 41.07 & 71.35 & 48.36 & \textbf{47.48} & 13.06 \\

\method{} (10\%) & 27.67 & 46.11 & 36.89 & 62.89 & 27.27 & 55.22 & 47.21 & 48.15 & 43.94 & 40.57 & 56.13 & 29.19 & 44.18 & 69.06 & 47.18 & 45.79 & 23.77 \\

\method{} (15\%) & 25.98 & 44.50 & 35.24 & 63.04 & 25.25 & 52.63 & 44.25 & 46.29 & 43.52 & 42.00 & 54.56 & 28.11 & 38.72 & 67.64 & 45.76 & 44.18 & 30.45\\

\bottomrule
\end{tabular}
    }
    \caption{The detailed performance of LLaMA3.1-8B trained on WizardLM data in terms of FLOPs and evaluation benchmarks for instruction following, knowledge, and reasoning. The datasets AE, TQA, HS, LQA, and WG correspond to AlpacaEval, TruthfulQA, HellaSwag, LogicQA, and WinoGrande, respectively.}
    \label{tab:wizardlm123}
\end{table*}

\section{Additional References for Heuristic-based Methods}
\label{sec:appendix_refs}

We list below all heuristic-based methods for automatic data selection in instruction tuning cited in Section~\ref{sec:related_work}:
~\citet{zhang2024recost},~\citet{Zhang2025TheBI}, ~\citet{Cao2023InstructionMH}, ~\citet{Chen2023AlpaGasusTA}, ~\citet{Liu2023WhatMG}, ~\citet{Bukharin2023DataDM}, ~\citet{Best}, ~\citet{Du2023MoDSMD}, ~\citet{Kung2023ActiveIT}, ~\citet{Li2023FromQT}, ~\citet{Li2024SuperfilteringWD}, ~\citet{Liu2024SelectITSI}, ~\citet{Wang2024HowDY}, ~\citet{Li2024SelectiveRS}, ~\citet{ge2024clustering}, ~\citet{he2024shed}, ~\citet{han2025automatic}, and ~\citet{wettig2024qurating}.

\end{document}